\documentclass{bmvc2k}
\usepackage{amssymb}
\usepackage{booktabs}
\usepackage{multirow}  
\usepackage{makecell}  
\usepackage{subfigure} 
\usepackage{float}  
\usepackage{pifont}
\usepackage{float}      

\usepackage{graphicx}   
\usepackage{placeins}

\title{GLip: A Global-Local Integrated Progressive Framework for Robust Visual Speech Recognition}

\addauthor{Tianyue Wang}{wangtianyue33@163.com}{1,2}
\addauthor{Shuang Yang}{shuang.yang@ict.ac.cn}{1,2}
\addauthor{Shiguang Shan}{sgshan@ict.ac.cn}{1,2}
\addauthor{Xilin Chen}{xlchen@ict.ac.cn}{1,2}

\addinstitution{
 University of Chinese Academy of  \\
 Sciences, \\
 Beijing, 100049, China
}

\addinstitution{
 State Key Laboratory of \\
 AI Safety,\\
 Institute of Computing Technology,\\
 Chinese Academy of Sciences,\\
 Beijing 100190, China
 }

\runninghead{WANG., YANG., SHAN., CHEN.}{GLip}

\begin{document}

\maketitle

\begin{abstract}
Visual speech recognition (VSR), also known as lip reading, is the task of recognizing speech from silent video. Despite significant advancements in VSR over recent decades, most existing methods pay limited attention to real-world visual challenges such as illumination variations, occlusions, blurring, and pose changes. To address these challenges, we propose GLip, a Global-Local Integrated Progressive framework designed for robust VSR. GLip is built upon two key insights: (i) learning an initial \textit{coarse} alignment between visual features across varying conditions and corresponding speech content facilitates the subsequent learning of \textit{precise} visual-to-speech mappings in challenging environments; (ii) under adverse conditions, certain local regions (e.g., non-occluded areas) often exhibit more discriminative cues for lip reading than global features. To this end, GLip introduces a dual-path feature extraction architecture that integrates both global and local features within a two-stage progressive learning framework. In the first stage, the model learns to align both global and local visual features with corresponding acoustic speech units using easily accessible audio-visual data, establishing a coarse yet semantically robust foundation. In the second stage, we introduce a Contextual Enhancement Module (CEM) to dynamically integrate local features with relevant global context across both spatial and temporal dimensions, refining the coarse representations into precise visual-speech mappings. Our framework uniquely exploits discriminative local regions through a progressive learning strategy, demonstrating enhanced robustness against various visual challenges and consistently outperforming existing methods on the LRS2 and LRS3 benchmarks. We further validate its effectiveness on a newly introduced challenging Mandarin dataset.
\end{abstract}

\section{Introduction}
\label{sec:intro}
\begin{figure}[H]
    \centering
    \begin{minipage}{0.48\textwidth}
        \centering
        \includegraphics[width=\textwidth]{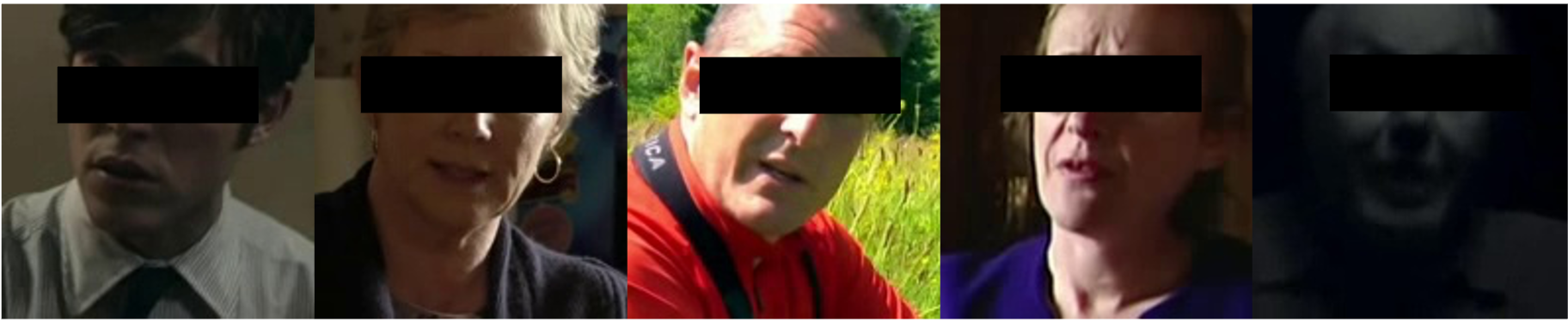}
        \\ 
        \small\textbf{(a) Illumination} 
        \label{fig:sub1}
    \end{minipage}
    \hspace{0.02\textwidth}
    \begin{minipage}{0.48\textwidth}
        \centering
        \includegraphics[width=\textwidth]{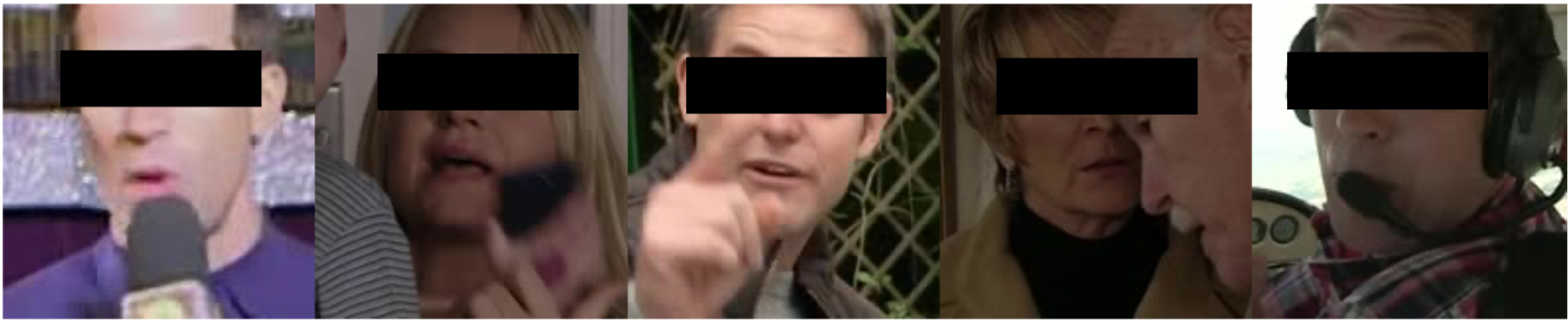}
        \\
        \small\textbf{(b) Occlusions}
        \label{fig:sub2}
    \end{minipage}
    \vspace{0.5em} 
    \begin{minipage}{0.48\textwidth}
        \centering
        \includegraphics[width=\textwidth]{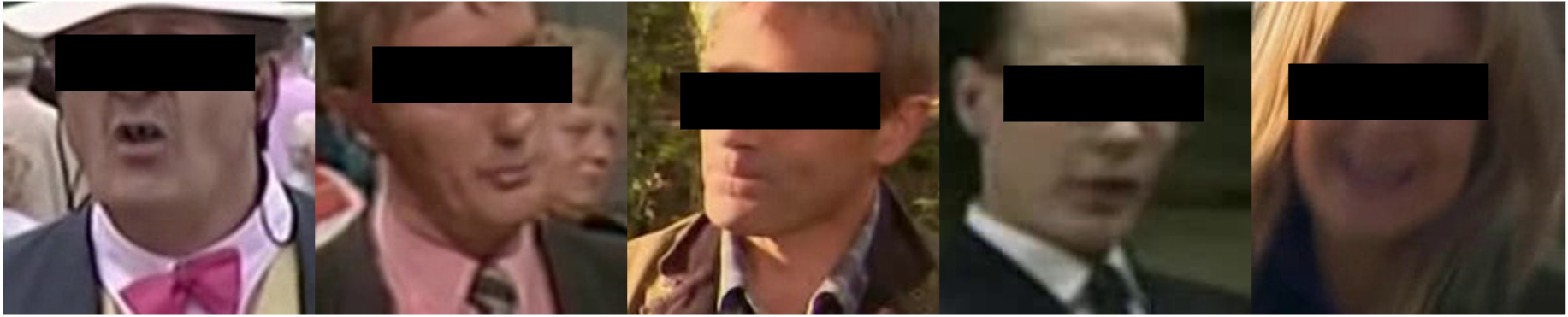}
        \\
        \small\textbf{(c) Motion blur}
        \label{fig:sub3}
    \end{minipage}
    \hspace{0.02\textwidth}
    \begin{minipage}{0.48\textwidth}
        \centering
        \includegraphics[width=\textwidth]{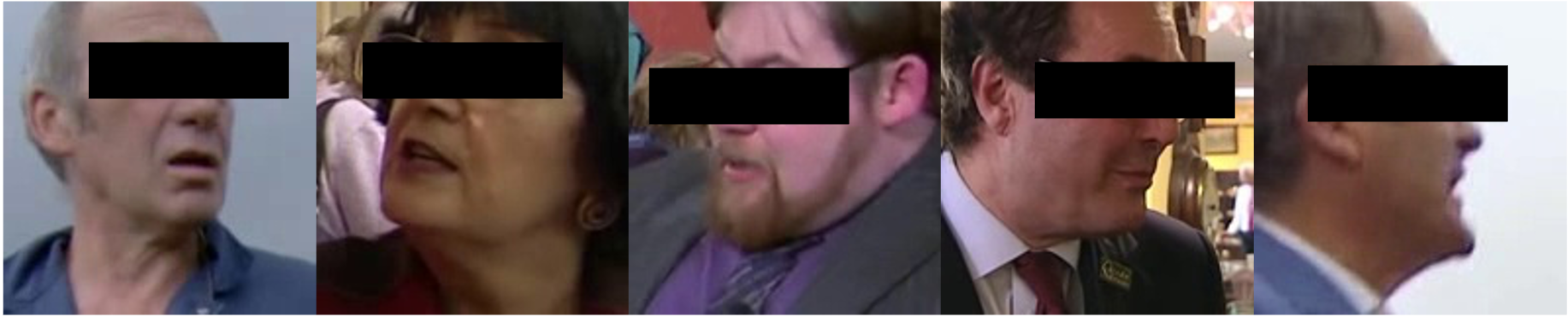}
        \\
        \small\textbf{(d) Head poses}
        \label{fig:sub4}
    \end{minipage}
    \caption{Examples from the public LRS2 dataset depicting various real-world conditions.}
    \label{fig:wer_factors}
\end{figure}

Visual speech recognition (VSR), or lip reading, involves recognizing speech from silent videos.
This task has received increasing attention due to its potential applications, such as aiding communication for aphasic patients, robust speech recognition in noisy environments, silent film dubbing, and security enhancement \cite{laux2023two,tye2007audiovisual,martinez2020lipreading,xu2020discriminative,haliassos2021lips}.
Despite significant advances in VSR, most existing methods lack emphasis on real-world challenges such as illumination variations, occlusions, blurring, and pose changes, as shown in Figure \ref{fig:wer_factors}. 



Existing efforts to address these challenges generally follow two directions. 
The first line of work focuses on collecting well-annotated data, such as the multi-view lip reading datasets OuluVS2 \cite{anina2015ouluvs2}, where multiple synchronized cameras capture the same utterance from different angles, enabling view-specific model training\cite{ma2021end, isobe2021multi} or pose-adaptive universal model development \cite{chung2017lip, maeda2021multi}.
The second line leverages data augmentation techniques to simulate visual degradation. Examples include 3DMM-based pose synthesis for multi-view lip reading \cite{cheng2020towards}, visual noise augmentations to improve noise robustness \cite{fernandez2023sparsevsr}, and synthetic data generation to simulate diverse poses and environmental conditions \cite{hao2025lipgen}.
While effective under certain conditions, these two lines face key limitations. Collecting annotated data of visual conditions is costly, while synthetic augmentations often struggle to bridge the domain gap between generated and real-world data. Furthermore, both tend to focus on specific and predefined types of degradation. 
All of these factors limit their ability to generalize across a wide range of real-world variations.

In this paper, we propose \textbf{GLip}, a Global-Local Integrated Progressive framework for robust lip reading under such challenging conditions. 
GLip is motivated by two key insights. 
First, learning a \textit{coarse} alignment between visual dynamics of various conditions and speech content facilitates subsequent learning of \textit{precise} visual-speech mappings in challenging scenarios.
Second, we recognize that in the presence of visual degradations, such as pose variations or occlusions, certain local regions (e.g., non-occluded areas) often exhibit more discriminative cues for accurate lip reading than other regions and even global features. 
To this end, GLip adopts a dual-path feature extraction architecture, progressively learned in two stages.
Given that different regions of a video simultaneously convey speech content despite varying visual quality, we introduce a local feature branch that identifies multiple informative regions, along with a global feature branch. We enforce both local and global features align with intrinsic speech information represented by acoustic speech units, establishing a coarse yet semantic correspondence between visual inputs across diverse conditions and the underlying speech content.
Subsequently, we introduce a Contextual Enhancement Module (CEM) to dynamically integrate local features with relevant global context in the second stage, refining the previously aligned representations into precise mappings from visual dynamics to speech content.

In summary, our main contributions are as follows:
\textbf{(1)} We propose GLip, a novel VSR framework that explicitly addresses real-world visual challenges by progressive learning with a dual-path feature extraction architecture.
\textbf{(2)} We provide a low-cost solution without expensive manually annotated or synthetic datasets. Instead, we leverage easily accessible unlabeled audio-visual data to learn a coarse alignment before refining to a precise visual-speech mapping.
\textbf{(3)} Extensive experiments on LRS2 and LRS3 demonstrate the strong generalization ability and robustness of our method in real-world scenarios.
\textbf{(4)} We contribute CAS-VSR-MOV20, a new challenging Mandarin VSR dataset for evaluation of VSR under real-world challenging conditions.

\section{Related Work}
\subsection{Visual Speech Recognition}
Deep learning has driven significant progress in VSR, with models based on spatio-temporal convolutions, attention mechanisms, and Transformers \cite{chung2017lip, afouras2018deep, martinez2020lipreading}, achieving remarkable success on benchmark datasets such as GRID \cite{cooke2006audio}, LRW \cite{yang2019lrw}, LRS2 \cite{afouras2018deep}, and LRS3 \cite{afouras2018lrs3}. However, most existing approaches place less emphasis on real-world challenges such as varying illumination, occlusions, and head pose changes, which significantly affect model robustness and generalization. Early attempts addressed pose variation via multi-view datasets like OuluVS2 \cite{anina2015ouluvs2}, requiring multiple synchronized cameras to construct view-specific feature extractors \cite{ma2021end, isobe2021multi} or pose-adaptive universal models \cite{chung2017lip,maeda2021multi}. While effective in controlled settings, such approaches suffer from limited scalability and practicality due to the high cost and complexity of data collection. Other studies have turned to data augmentation techniques, including 3DMM-based pose synthesis \cite{cheng2020towards}, visual noise augmentations \cite{fernandez2023sparsevsr}, and synthetic data generation \cite{hao2025lipgen}. Nevertheless, these methods often suffer from domain gaps between synthetic and real-world data, leading to overfitting and poor generalization, as observed in \cite{hao2025lipgen}. Moreover, most approaches are designed to handle specific, predefined types of degradation, failing to comprehensively address the diverse variations encountered in practice. In contrast, our method leverages easily obtainable unlabeled audio-visual data to progressively learn robust visual representations and dual-path integration to capture informative cues against various types of visual challenges.

\subsection{Robust Multi-modal Speech Recognition}
Both Automatic Speech Recognition (ASR) and Visual Speech Recognition (VSR) face considerable challenges under adverse conditions, such as background noise, visual distortions, and temporal misalignment. To address these issues, multi-modal approaches have been proposed to leverage the complementarity of audio and visual cues. Most existing studies aim to enhance ASR performance using visual inputs in noisy acoustic environments \cite{shi2022robust,hu2023hearing,chen2023leveraging,kim2024learning,ithal2024enhancing,hu2023cross}, while fewer have addressed the challenge of degraded visual inputs for VSR. Recent works have begun addressing both audio and visual degradation in audio-visual speech recognition, including introducing synthetic noise and occlusions to simulate adverse conditions \cite{hong2023watch}, 
employing image restoration techniques using generative models to recover occluded inputs \cite{wang2024restoring}, and leveraging cross-modal reconstruction methods to compensate for corrupted visual features using audio cues \cite{kim2025multi}.
However, robust VSR under diverse visual degradations remains largely underexplored. To the best of our knowledge, our work is the first to explicitly improve VSR performance under a variety of visual challenges by leveraging auxiliary audio information during training.

\section{Method}
\begin{figure}[H]
  \centering
  \includegraphics[scale=0.28]{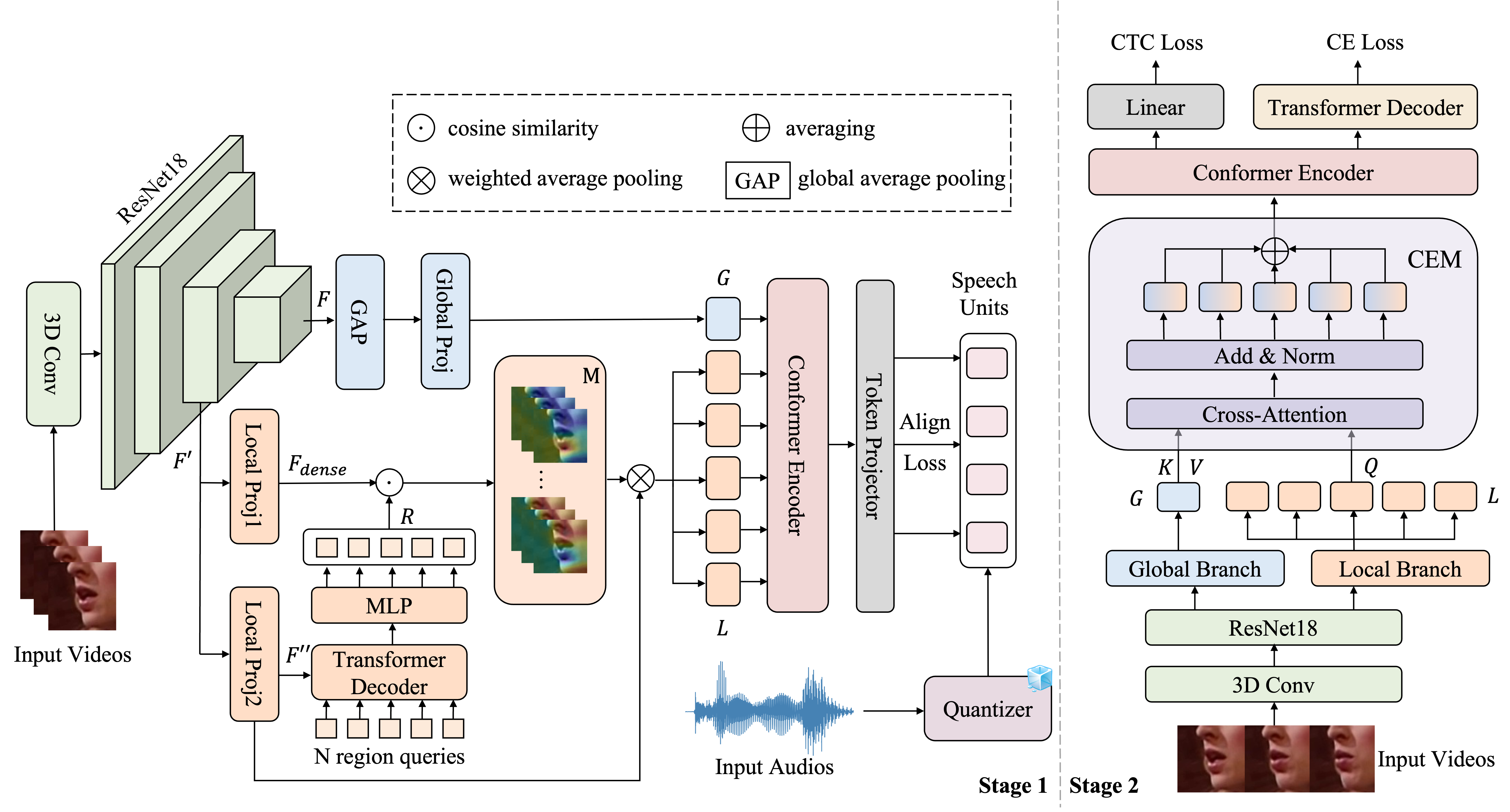}
\vspace{-5pt} 
  \caption{Overview of the proposed GLip.}
  \label{fig:glip}
\vspace{-5pt}
\end{figure}

The proposed GLip adopts a dual-path feature extraction architecture within a progressive learning framework, as illustrated in Figure \ref{fig:glip}. Our framework consists of two stages: (1) Coarse global-local audio-visual alignment that establishes robust initial representations, and (2) Context-aware refinement of visual-speech mapping that integrates local and global information to refine the coarse alignment into precise visual-speech mappings.

\subsection{Stage 1: Coarse Global-Local Audio-Visual Alignment}
The first stage aims to establish a coarse yet semantically aligned mapping between global and local visual features and corresponding speech units. The underlying idea is that different regions simultaneously convey the same speech content despite varying visual conditions. By enforcing consistency among these features with respect to the shared speech content, we build a robust initialization for subsequent refinement.

\noindent\textbf{Dual-Path Feature Extraction Architecture.}
Given an input video sequence $x$, it is processed by a visual front-end consisting of a 3D CNN  followed by ResNet-18. 
We extract feature maps from the final layer and penultimate layer of ResNet-18, denoted as $\mathit{F}\in \mathbb{R}^{T \times C \times H \times W}$ and $\mathit{F'} \in \mathbb{R}^{T \times c \times h \times w}$ respectively.
\textit{Global Feature Branch.} To obtain global visual representations, we apply global average pooling (GAP) over the spatial dimensions of \(\mathit{F}\) to get a latent representation, which is then passed through a global projector to obtain global embeddings $\mathit{G} \in \mathbb{R}^{T \times D}$, where $D$ is the embedding dimension.
\textit{Local Feature Branch.} In parallel, we use $\mathit{F'}$, which preserves more spatial detail than $\mathit{F}$, as input to two separate projectors. The first produces a dense feature map $\mathit{F_{\text{dense}}} \in \mathbb{R}^{T \times D \times h \times w}$, and the second outputs an intermediate representation $\mathit{F''} \in \mathbb{R}^{T \times D \times h \times w}$ for subsequent attention operations.
To localize discriminative regions, we employ a Transformer decoder followed by a MLP. The decoder takes $N$ learnable region queries as input, which act as trainable tokens that dynamically attend to informative areas across the feature map $\mathit{F''}$, enabling the model to automatically identify multiple regions relevant to speech without explicit annotations.
The decoder outputs region-specific embeddings $\mathit{R} \in \mathbb{R}^{N \times T \times D}$, where each vector corresponds to one identified region across time.
To measure the correspondence between each region and spatial locations in the feature map, we compute the cosine similarity between $\mathit{R}$ and $\mathit{F_{\text{dense}}}$ across the channel dimension, producing soft clustering assignments $\mathit{S} \in \mathbb{R}^{N \times T \times h \times w}$.
After applying a softmax operation over the channel dimension, we obtain attention maps $\mathit{M}$ which highlight the most salient spatial locations per region.
Finally, we obtain local embeddings via weighted average pooling:
\begin{equation}
\mathit{l_{t,n}} = \frac{\sum_{u,v} \mathit{M}[t, n, u, v] \mathit{F''}[t, :, u, v]}{\sum_{u,v} \mathit{M}[t, n, u, v]},
\end{equation}
where $\mathit{l_{t,n}} \in \mathbb{R}^D$ is the local embedding for the $n$-th local region at frame $t$. We then concatenate all $N$ region features per frame for the final local embedding $\mathit{L} \in \mathbb{R}^{T \times N \times D}$.

\noindent\textbf{Audio-Visual Alignment Objective.}
Both the global and local embeddings are subsequently fed into a Conformer encoder and a token projector to predict quantized speech units derived from vq-wav2vec \cite{baevski2019vq}, denoted as \( z = \{z_t\}_{t=1}^T \). 
We apply cross-entropy loss to predict \( z_t \) from input video frames,  where each video frame corresponds to four speech units. \( p(z_t | x) \) denotes the model’s output at time \( t \) based on the video \( x \). The alignment loss is computed separately for the local and global branches.
The total alignment loss is:
\begin{equation}
    \mathcal{L}_{\text{align}} = \mathcal{L}_{\text{global}} + \mathcal{L}_{\text{local}},
\end{equation}
where
\begin{equation}
    \mathcal{L}_{\text{global}} = -\frac{1}{T} \sum_{t=1}^T \log p(z_t | \mathit{G}), 
\end{equation}
\begin{equation}
\mathcal{L}_{\text{local}}  = -\frac{1}{NT} \sum_{n=1}^N \sum_{t=1}^T \log p(z_t | \mathit{L}_n).
\end{equation}

\subsection{Stage 2: Context-Aware Refinement of Visual-Speech Mapping}
The second stage builds on the pretrained visual front-end and dual-path feature extraction architecture from the first stage. While the first stage provides a strong initial alignment, it lacks direct interaction between global and local features. Global features offer semantic completeness but limited spatial detail, whereas local features provide spatial precision but are contextually isolated.
To leverage their complementary strengths, we introduce a Contextual Enhancement Module (CEM) based on a cross-attention mechanism to facilitate effective interaction between global and local representations across both spatial and temporal dimensions.

\noindent\textbf{Contextual Enhancement Module.} 
Given a lip movement video $x$, we obtain global embeddings $\mathit{G} \in \mathbb{R}^{T \times D}$ from the global branch and $N$ local embeddings $\{\mathit{L}_n \in \mathbb{R}^{T \times D}\}_{n=1}^N$ from the local branch. 
Each local embedding sequence $\mathit{L}$ serves as a query to attend to the global sequence. This design allows each local region at every time step to integrate the most relevant global context, thereby capturing not only spatial dependencies across local regions but also temporal coherence across video frames. 
The interaction is implemented using a multi-layer cross-attention mechanism. For each layer  (\(k=1...K\)), we update each local embedding \( \mathit{L}_n^{(k-1)} \) as follows:

\begin{equation}
    \mathit{G}'^{(k-1)}=\text{Softmax}\left(
    \frac{(\mathit{L}_{n}^{(k-1)} \mathit{W}_{Q}^{(k-1)})(\mathit{G} \mathit{W}_{K}^{(k-1)})^\top}
    {\sqrt{D_k}} \right) (\mathit{G} \mathit{W}_{V}^{(k-1)})
\end{equation}

\begin{equation}
    \mathit{L}_{n}^{(k)} = \text{LayerNorm}\left(\mathit{L}_{n}^{(k-1)} + \mathit{G}'^{(k-1)}\mathit{W}_O^{(k-1)}\right)
\end{equation}
where $\mathit{W}_{Q},\mathit{W}_{K},\mathit{W}_{V}, \mathit{W}_O \in \mathbb{R}^{D \times D}$. Starting with $\mathit{L}_{n}^{(0)} = \mathit{L}_n$, this process iteratively enhances local features with global context across $K$ layers.
The enhanced local features \(\{\mathit{L}_{n}^{(K)}\}_{n=1}^N \in \mathbb{R}^{T \times D}\) are aggregated by averaging across regions:  
\begin{equation}
    \bar{\mathit{L}}^{(K)} = \frac{1}{N}\sum_{n=1}^N \mathit{L}_{n}^{(K)} \in \mathbb{R}^{T \times D}
\end{equation}
\(\bar{\mathit{L}}^{(K)}\) is then processed by a Conformer encoder to capture temporal dynamics and a Transformer decoder autoregressively to generate the text sequence.

\noindent\textbf{Hybrid CTC-Attention loss.}
The second stage employs the hybrid CTC-attention loss for training.
The total training objective combines both losses with a balancing weight $\lambda \in [0,1]$:
\begin{equation}
    \mathcal{L}_{\text{total}} = \lambda \mathcal{L}_{\text{CTC}} + (1 - \lambda)\mathcal{L}_{\text{CE}}.
\end{equation}
Here, $\mathcal{L}_{\text{CTC}}$ is the CTC loss \cite{graves2006connectionist}, which enforces frame-wise predictions under the conditional independence assumption. $\mathcal{L}_{\text{CE}}$ denotes the cross-entropy loss, which predicts the next label conditioned on the previous outputs and the input sequence.

\section{Experiments}
\subsection{Datasets}
We conduct experiments on two widely used lip reading datasets: LRS2 ~\cite{afouras2018deep} and LRS3 ~\cite{afouras2018lrs3}.
LRS2 is a large-scale audio-visual dataset collected from BBC programmes, which contains approximately 224.5 hours of video data. It consists of sentences spoken by a diverse set of speakers in various linguistic contexts.
LRS3 is a dataset collected from TED talks, encompassing over 5.5 thousand unique speakers with around 439 hours of data. It covers a wide range of topics and speaking styles, providing a rich and natural source of visual speech data.
Both datasets reflect challenging real-world visual conditions, including significant head pose variations, varying illumination, partial occlusions (e.g., hands or objects covering the mouth), and motion blur caused by fast facial or camera movements. These characteristics make LRS2 and LRS3 highly representative benchmarks for evaluating the performance and robustness of visual speech recognition systems.


\subsection{Implementation Details}  
Following common practices ~\cite{martinez2020lipreading,ma2022visual,ma2023auto}, we crop the lip region to a fixed bounding box of size 96×96 pixels. Quantized speech units are extracted using vq-wav2vec ~\cite{baevski2019vq}.  
During training, we apply a series of data augmentation techniques, including random cropping to 88×88 pixels, adaptive time masking, and random image degradation, to enhance model robustness and generalization.  
For the backbone architecture, we use a Conformer encoder with 12 layers, and a Transformer decoder with 6 layers, both employing attention dimensions of 768.  
The number of heatmaps $N$ is empirically set to 5. 
Based on experimental results, we adopt a single-layer Transformer decoder in the local feature branch, with an attention dimension of 768. The CEM follows the same configuration.
We optimize the model using AdamW ~\cite{loshchilov2017decoupled} with momentum $\beta_1 = 0.9$, $\beta_2 = 0.98$. The value of $\lambda$ is set to 0.1.

\subsection{Comparison with Others}
We compare GLip with a variety of state-of-the-art VSR methods on the LRS2 and LRS3 benchmarks. The results are summarized in Table~\ref{tab:sota}, measured using Word Error Rate (WER). 
GLip achieves the best performance on both datasets with a WER of 28.1\% on LRS2 and 30.1\% on LRS3. 
Furthermore, when using the audio-visual data in LRS3 for the first stage, GLip further reduces the WER on LRS2 to 27.4\%, demonstrating the potential of leveraging widely available audio-visual pairs to improve performance. 
Notably, while using the standard 96 × 96 lip region input  (in line with most existing approaches), GLip even outperforms methods that utilize larger input scales, demonstrating its remarkable effectiveness in exploiting discriminative visual cues from constrained visual regions.
\begin{table}[H]
\centering
\caption{Comparison of different methods on LRS2 and LRS3 datasets.}
\label{tab:sota}
\begin{tabular}{lcccccc}
\toprule
\multirow{2}{*}{\textbf{Method}} & \multirow{2}{*}{\textbf{Input Scale}} & \multirow{2}{*}{\textbf{Unlab hours}} & \multirow{2}{*}{\textbf{Lab hours}} & \multicolumn{2}{c}{\textbf{WER}} \\
\cmidrule(lr){5-6}
 & & & & \textbf{LRS2} & \textbf{LRS3} \\
\midrule
MV-WAS \cite{chung2017lip}       & 224 × 224    & --        & 223       & 70.4  & --   \\
TDNN \cite{yu2020audio}         & 112 × 112    & --        & 223       & 48.9  & --   \\
CM-Seq2Seq \cite{ma2021end}   & 96 × 96      & --   & 223/438       & 39.1  & 46.9 \\
CM-Aux \cite{ma2022visual}       & 96 × 96      & --        & 223/438      & 32.9  & 37.9 \\
LiRA \cite{ma2021lira}         & 96 × 96      & 438       & 223       & 38.8  & --   \\
RAVEn \cite{haliassos2022jointly}        & 96 × 96      & 433       & 223/433       & 32.1  & 39.1 \\
AutoAVSR \cite{ma2023auto}     & 96 × 96      & --       & 438       & --  & 36.3 \\
\text{ES}$^{3}$ \cite{zhang2024es3} & 96 × 96 & 433 & 223/433 & 28.7  & 37.9 \\
SyncVSR \cite{ahn2024syncvsr}      & 128 × 128    & --    & 223/438       & 28.9  & 31.2 \\
\textbf{GLip (Ours)} & 96 × 96     & 223   & 223       & \textbf{28.1} & -- \\
\textbf{GLip (Ours)} & 96 × 96     & 433   & 223/433       & \textbf{27.4} & \textbf{30.1} \\
\bottomrule
\end{tabular}
\end{table}

\subsection{Ablation Study}
\subsubsection{Quantitative Analysis}
To evaluate the effectiveness of key components in our proposed GLip framework, we conduct a series of ablation studies on the LRS2 dataset. The results are summarized in Table\ref{tab:ablation_study}.

\noindent\textbf{Effectiveness of Progressive Learning.}
We first evaluate the necessity of progressive training. 
When progressively initialize the training with global features only, The global branch with cross-modal alignment in the first stage achieves a WER of 28.45\%, representing a significant 5.2\% relative improvement over the baseline \cite{ma2023auto}. This result confirms the effectiveness of coarse alignment as a foundation for subsequent refinement.

\noindent\textbf{Effectiveness of Dual-Path Feature Extraction Architecture.}
We investigate the contribution of the dual-path feature extraction architecture. We introduce the local branch alongside the global branch in the first stage. The resulting WER is further reduced to 28.24\%, suggesting that local information provides complementary cues to the global representation, even in early training.
However, when we replace the global branch with only the local branch in the second stage,  the performance degrades to 28.30\%. This indicates that relying solely on local features may lack sufficient contextual understanding, validating the importance of leveraging both global and local information through our dual-path design.

\noindent\textbf{Effectiveness of CEM.}
We compare different feature fusion strategies. We simply average the global and local features (AVG) in the second stage, which leads to a worse WER of 28.94\%. In contrast, our proposed CEM achieves the best WER of 28.10\%. This demonstrates the advantage of adaptive, context-aware fusion over naive combination.

\begin{table}[H]
\centering
\caption{Ablation Study on LRS2.}
\label{tab:ablation_study}
\begin{tabular}{ccccc}
\toprule
\multirow{2}{*}{\textbf{Method}} & \multicolumn{2}{c}{\textbf{Branches}} & \multirow{2}{*}{\textbf{Fusion}} & \multirow{2}{*}{\textbf{WER}} \\
\cmidrule(lr){2-3}
& \textbf{Stage1} & \textbf{Stage2} & & \\
\midrule
Baseline & \ding{53} & Global & \ding{53} & 30.01 \\
\midrule
\multirow{6}{*}{Ours} 
& Global & Global & \ding{53} & 28.45 \\
& Global\&Local & Global & \ding{53} & 28.24 \\
& Global\&Local & Local & \ding{53} & 28.30 \\
& Global\&Local & Global\&Local & AVG & 28.94 \\
& Global\&Local & Global\&Local & CEM & \textbf{28.10} \\
\bottomrule
\end{tabular}
\end{table}

\subsubsection{Qualitative Analysis}
\begin{figure}[H]
    \centering
    \begin{minipage}{0.38\textwidth}
        \centering
        \includegraphics[width=\textwidth]{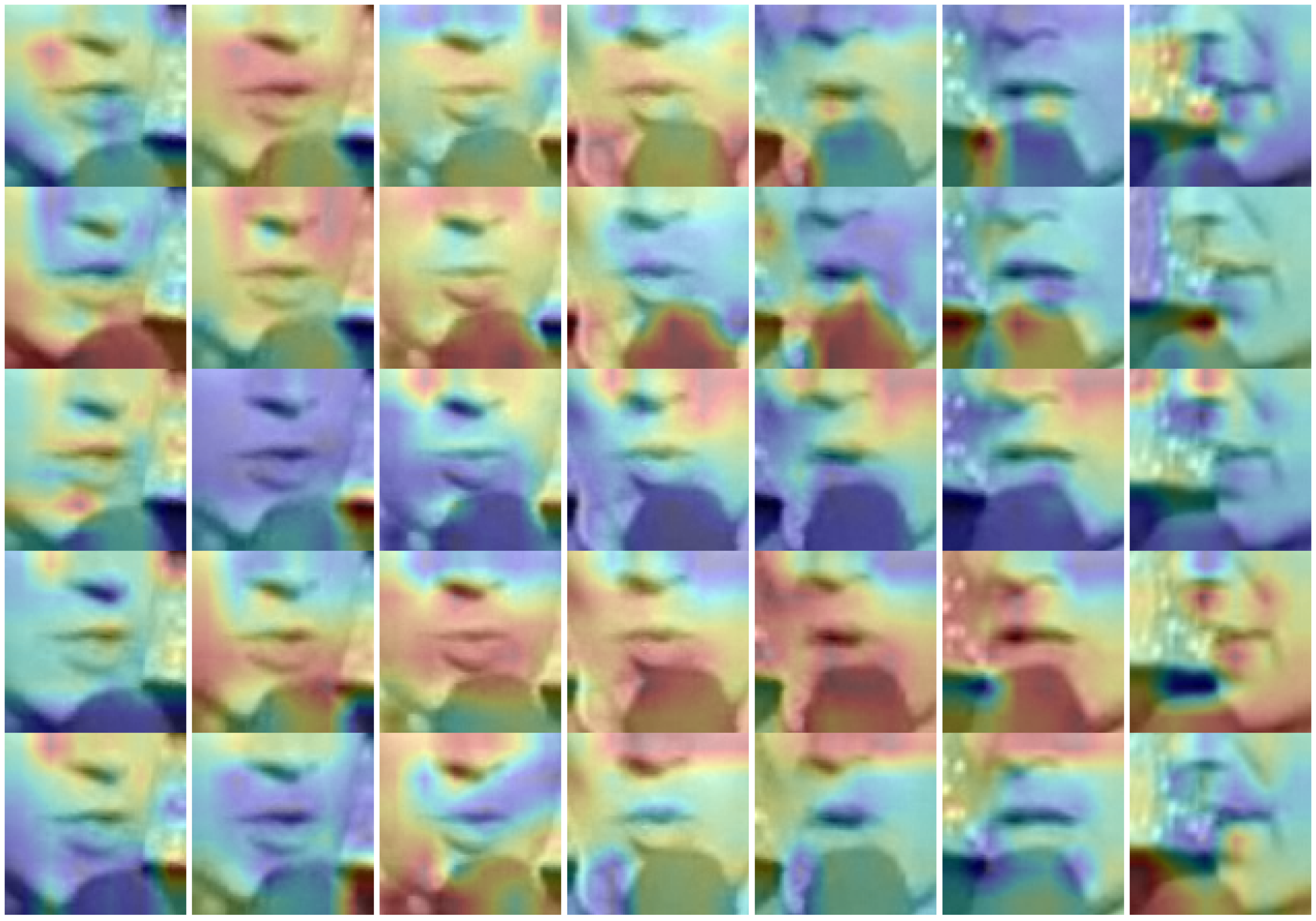}
        \\ 
        \small\textbf{(a) Stage 1} 
        \label{fig:sub1}
    \end{minipage}
    \hspace{0.016\textwidth}
    \begin{minipage}{0.38\textwidth}
        \centering
        \includegraphics[width=\textwidth]{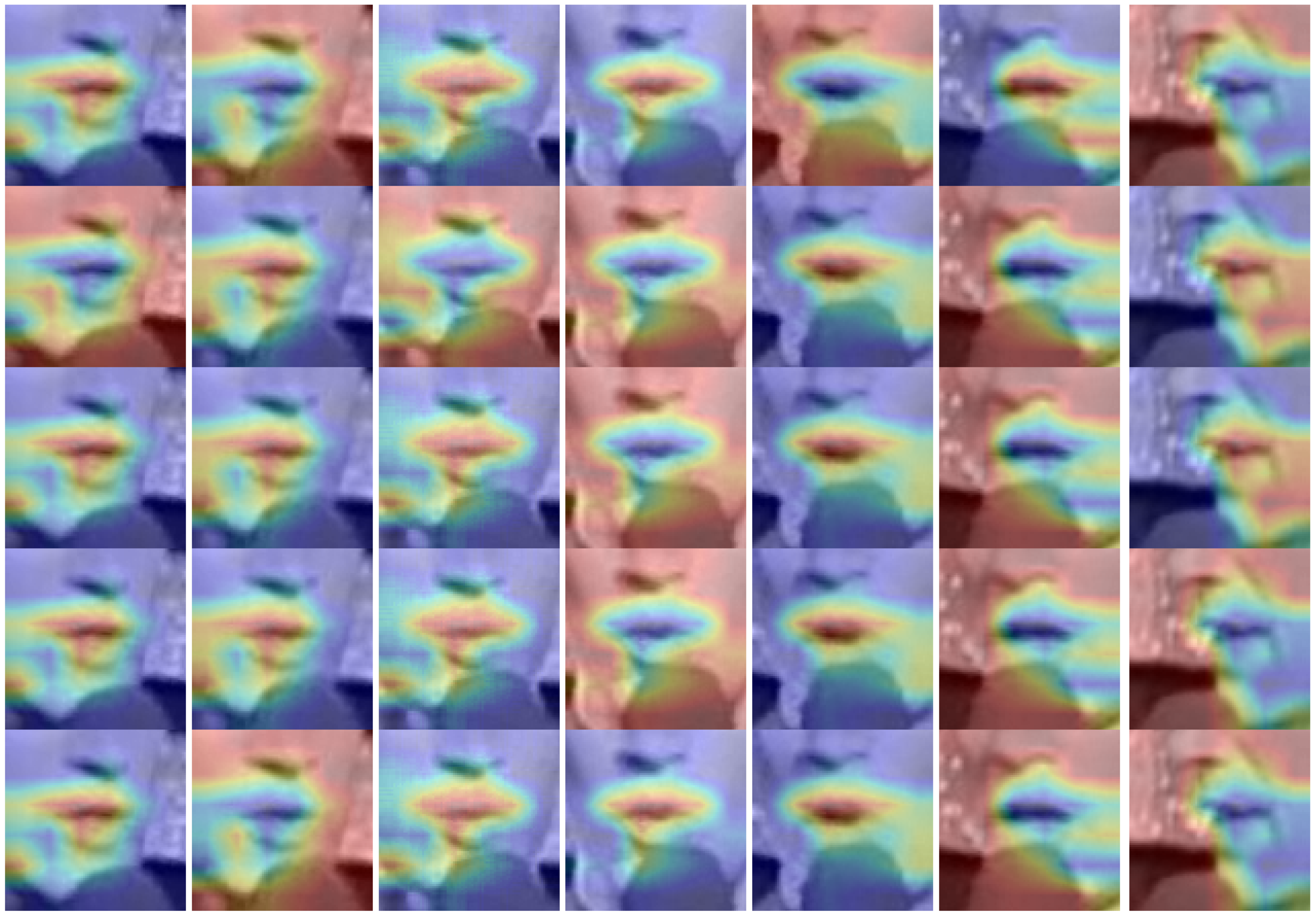}
        \\ 
        \small\textbf{(b) Stage 2} 
        \label{fig:sub2}
    \end{minipage}
    \caption{Visualization of heatmaps from the local feature branches in Stage 1 and Stage 2. The horizontal and vertical axes represent time steps and different regions, respectively.}
    \label{fig:heatmaps}
    \vspace{-7pt} 
\end{figure}

To better understand the behavior of GLip, we visualize heatmaps generated by the local feature branch in both training stages. Figure~\ref{fig:heatmaps} presents a representative example from the LRS2 dataset under challenging conditions including large pose variations and microphone occlusions.
In Stage 1, the heatmaps primarily highlight regions around the lips, nose, jaw, and cheek muscles. This indicates that these areas contain crucial local visual cues for modeling speech-related dynamics. However, the attention is relatively diffused and lacks fine-grained localization.
By contrast, in Stage 2—where global and local features are integrated via the CEM module—the heatmaps exhibit a much sharper focus on the lip region even under pose variations and partial occlusions. Moreover, the regions incorporate richer contextual information, benefiting from the global features that provide spatial and temporal context. 

\subsection{Performance under Diverse Visual Conditions}
We conduct a detailed analysis on LRS2 and LRS3 by categorizing the test sets based on varying degrees of four common challenging visual conditions: illumination, occlusion, blur, and pose variation.
We will release this categorized test set partition to foster further research and development in the field\textsuperscript{1}.
\footnotetext[1]{\url{https://github.com/Physicsmile/GLip}.} 
As no publicly available models have been trained exclusively on LRS2, we reproduce two recent methods—AutoAVSR~\cite{ma2023auto} and SyncVSR~\cite{ahn2024syncvsr}—to enable direct comparison with GLip.
For LRS3, we use the official checkpoints of RAVEn~\cite{haliassos2022jointly} and AutoAVSR~\cite{ma2023auto}, ensuring fair comparisons under consistent experimental settings. The detailed results are summarized in Table~\ref{tab:comparison_lrs_combined}.

\noindent\textbf{Illumination.} We categorize illumination into bright (B), moderate (M), and dark (D). GLip consistently outperforms other methods under all lighting conditions, with especially strong gains in bright and dark settings, where visual signals are degraded. For example, GLip achieves 32.05\% WER on LRS3, surpassing AutoAVSR and RAVEn by 5\% and 16.77\%, respectively, demonstrating its robustness to illumination variation.

\noindent\textbf{Occlusion.} We consider two occlusion states: non-occluded (N) and occluded (Y).
Occlusions—such as hands or objects partially covering the lower face, lip motion cues are partially blocked, leading to substantial performance drops for conventional models. 
However, GLip still achieves a WER of 35.10\% on LRS3 under occlusion, outperforming AutoAVSR (40.99\%) and RAVEn (41.77\%). This highlights the effectiveness of GLip’s local feature branch in utilizing the remaining visible discriminative regions, complemented by contextual cues from the global branch. A similar trend is observed in the LRS2 results.

\noindent\textbf{Blur.}
Blur is categorized as clear (C), medium (M), and blurry (B). As blurriness increases, the lip contours and local textures become progressively less distinct. Under the blurry condition on LRS3, GLip achieves a WER of 44.15\%, outperforming AutoAVSR (47.25\%) and RAVEn (52.34\%). This superior performance can be largely attributed to GLip's ability to extract stable structural cues and effectively incorporate contextual information through its progressive learning strategy.

\noindent\textbf{Pose.} Pose variation is defined based on yaw angle into three categories: slight (S: 0°–30°), medium (M: 30°–60°), and large (L: 60°–90°). Larger angles introduce self-occlusion and substantial visual distortion. GLip demonstrates strong adaptability across all pose levels. It achieves a WER of 31.19\% under large pose variation on LRS3, outperforming AutoAVSR (37.61\%) and RAVEn (44.04\%). 
This robustness stems from the global branch's spatial continuity modeling and the local branch's precise focus on visible lip regions, ensuring stable performance despite extreme viewpoint shifts. 
\begin{table}[H]
\centering
\small
\caption{Comparison of WER(\%) on LRS2 and LRS3 under different visual conditions.}
\label{tab:comparison_lrs_combined}
\begin{tabular}{>{\centering\arraybackslash}p{0.8cm} >{\centering\arraybackslash}p{1.9cm} *{11}{>{\centering\arraybackslash}p{0.4cm}}}
\toprule
\multirow{2}{*}{\textbf{Dataset}} & \multirow{2}{*}{\textbf{Method}} & 
\multicolumn{3}{c}{\textbf{Illumination}} & 
\multicolumn{2}{c}{\textbf{Occlusion}} & 
\multicolumn{3}{c}{\textbf{Blur}} & 
\multicolumn{3}{c}{\textbf{Pose}} \\
\cmidrule(lr){3-5} \cmidrule(lr){6-7} \cmidrule(lr){8-10} \cmidrule(lr){11-13}
& & \textbf{B} & \textbf{M} & \textbf{D} & 
\textbf{N} & \textbf{Y} & 
\textbf{C} & \textbf{M} & \textbf{B} & 
\textbf{S} & \textbf{M} & \textbf{L} \\
\midrule
\multirow{3}{*}{LRS3} 
& RAVEn\cite{haliassos2022jointly} & 48.82 & 37.97 & 41.79 & 38.52 & 41.77 & 32.84 & 40.61 & 52.34 & 37.42 & 42.77 & 44.04 \\
& AutoAVSR\cite{ma2023auto} & 37.05 & 34.25 & 37.50 & 34.01 & 40.99 & 30.21 & 35.03 & 47.25 & 33.22 & 39.28 & 37.61 \\
& GLip (Ours) & \textbf{32.05} & \textbf{29.35} & \textbf{33.50} & \textbf{30.51} & \textbf{35.10} & \textbf{24.64} & \textbf{30.53} & \textbf{44.15} & \textbf{28.79} & \textbf{33.63} & \textbf{31.19} \\
\midrule
\multirow{3}{*}{LRS2} 
& AutoAVSR\cite{ma2023auto} & 30.44 & 28.33 & 34.70 & 29.48 & 38.68 & 30.16 & 29.64 & 32.89 & 29.51 & 29.42 & 49.16 \\
& SyncVSR\cite{ahn2024syncvsr} & 28.74 & 26.87 & 33.87 & 28.07 & 38.68 & 28.05 & 28.89 & 31.91 & 28.08 & 28.30 & 47.49 \\
& GLip (Ours) & \textbf{28.40} & \textbf{26.82} & \textbf{31.69} & \textbf{27.58} & \textbf{36.84} & \textbf{27.98} & \textbf{27.96} & \textbf{30.92} & \textbf{27.46} & \textbf{28.04} & \textbf{44.13} \\
\bottomrule
\end{tabular}
\vspace{-10pt} 
\end{table}

\subsection{Results on CAS-VSR-MOV20}
Furthermore, we introduce a new challenging dataset, CAS-VSR-MOV20\textsuperscript{2}, for evaluation. This dataset comprises short video clips (up to 3 minutes in length) sourced from 20 Chinese movies available on public platforms such as YouTube. It covers a variety of visual conditions, including diverse lighting environments, occlusions, blurring and pose variations.
\footnotetext[2]{\url{https://github.com/VIPL-Audio-Visual-Speech-Understanding/CAS-VSR-MOV20}.}
\begin{figure}[H]
\centering
\includegraphics[width=0.7\textwidth]{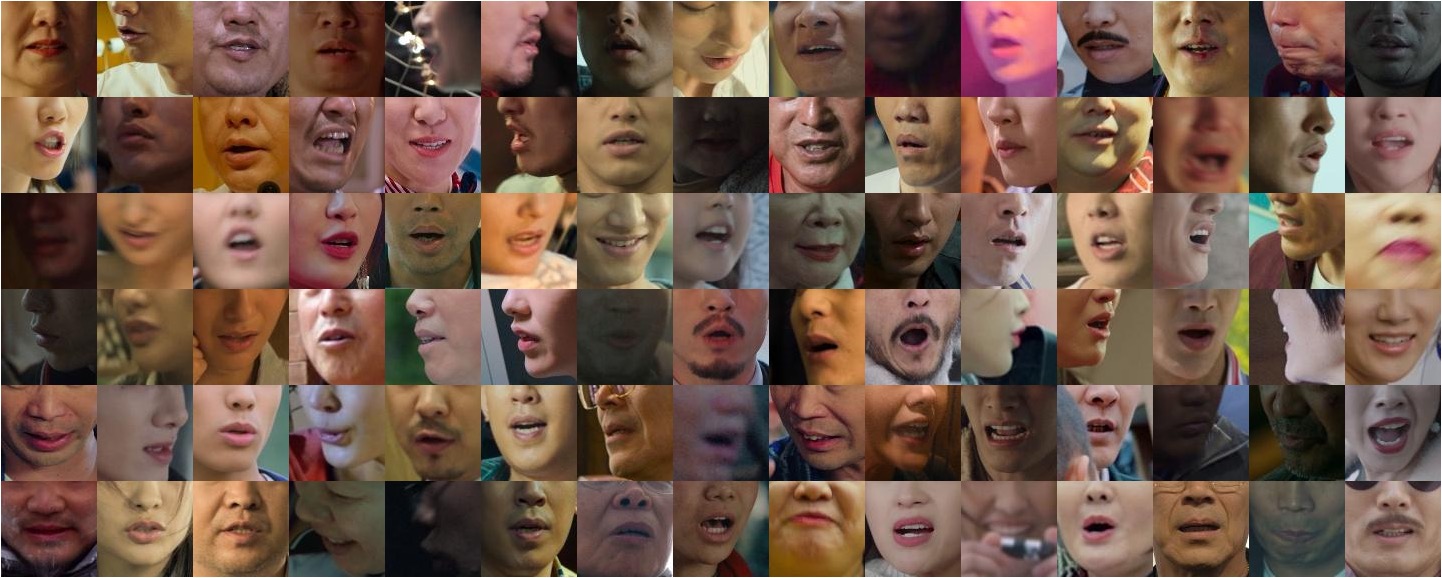}
\vspace{-5pt} 
\caption{Examples from CAS-VSR-MOV20.}
\label{fig:mov20}
\vspace{-5pt} 
\end{figure}
We conducted experiments under two settings based on CAS-VSR-S101\cite{zhang2024es3}: training from scratch on CAS-VSR-S101, and loading the dual-path feature extractor pre-trained on LRS3 in Stage 1 before training on CAS-VSR-S101. All experiments were evaluated on the validation and test sets of CAS-VSR-MOV20. The results are presented in Table \ref{tab:mov20}.

As shown, the baseline model achieves only 93.05\% CER on the validation set and 91.73\% on the test set, demonstrating the highly challenging nature of CAS-VSR-MOV20. In contrast, GLip brings substantial improvements, reducing the CER by 5.01\% and 3.95\% on the two subsets respectively, further validating its effectiveness and generalizability under different visual challenges. Moreover, initializing with pre-trained weights from LRS3 yields additional performance gains, despite the linguistic differences between the datasets.

\begin{table}[H]
\centering
\caption{Results on CAS-VSR-MOV20 dataset.}
\label{tab:mov20}
\begin{tabular}{@{}cccc@{}}
\toprule
\multirow{2}{*}{Method} & \multirow{2}{*}{Setting} & \multicolumn{2}{c}{CER(\%)} \\
\cmidrule(l){3-4}
 & & MOV20-Val & MOV20-Test \\
\midrule
Baseline & From Scratch & 93.05 & 91.73 \\
GLip & From Scratch & 88.04 & 87.78 \\
GLip & Load Pre-trained Model & 85.64 & 84.72 \\
\bottomrule
\end{tabular}
\end{table}


\section{Conclusion}
We propose GLip, a novel Global-Local Integrated Progressive framework for robust visual speech recognition. 
By progressively learning coarse-to-precise visual-speech mappings and adaptively integrating global and local features, GLip effectively addresses diverse visual challenges such as illumination variations, occlusions, blurring, and pose changes. Experiments on two popular English VSR datasets LRS2 and LRS3, as well as the newly released challenging Mandarin dataset CAS-VSR-MOV20, demonstrate appealing performance, highlighting its strong generalization capability for real-world applications.

\section{Acknowledgements}
This work is partially supported by National Natural Science Foundation of China (No. U24A20332, 62276247).

\bibliography{egbib}
\end{document}